# Human-Like Decision Making for Autonomous Driving: A Noncooperative Game Theoretic Approach

Peng Hang, Chen Lv, *Senior Member, IEEE*, Yang Xing, Chao Huang, and Zhongxu Hu

*Abstract*—Considering that human-driven vehicles and autonomous vehicles (AVs) will coexist on roads in the future for a long time, how to merge AVs into human drivers' traffic ecology and minimize the effect of AVs and their misfit with human drivers, are issues worthy of consideration. Moreover, different passengers have different needs for AVs, thus, how to provide personalized choices for different passengers is another issue for AVs. Therefore, a human-like decision making framework is designed for AVs in this paper. Different driving styles and social interaction characteristics are formulated for AVs regarding driving safety, ride comfort and travel efficiency, which are considered in the modeling process of decision making. Then, Nash equilibrium and Stackelberg game theory are applied to the noncooperative decision making. In addition, potential field method and model predictive control (MPC) are combined to deal with the motion prediction and planning for AVs, which provides predicted motion information for the decision-making module. Finally, two typical testing scenarios of lane change, i.e., merging and overtaking, are carried out to evaluate the feasibility and effectiveness of the proposed decision-making framework considering different human-like behaviors. Testing results indicate that both the two game theoretic approaches can provide reasonable human-like decision making for AVs. Compared with the Nash equilibrium approach, under the normal driving style, the cost value of decision making using the Stackelberg game theoretic approach is reduced by over 20%.

## I. Introduction

DECISION making is a vital part of autonomous driving technology. According to the information provided by the environment perception module, proper driving behaviors are planned by the decision-making module and sent to the motion control module [1]. Therefore, decision making is usually regarded as the brain of AV and bridges environment perception and motion control. The driving performance of AVs is effectively affected by the decision-making module, including driving safety, ride comfort, travel efficiency, and energy consumption [2].

This work was supported in part by the SUG-NAP Grant (No. M4082268.050) of Nanyang Technological University, Singapore, and A*STAR Grant (No. 1922500046), Singapore.

P. Hang, C. Lv, Y. Xing, C. Huang and Z. Hu are with the School of Mechanical and Aerospace Engineering, Nanyang Technological University, Singapore 639798. (E-mails: {peng.hang, lyuchen, xing.yang, chao.huang, zhongxu.hu}@ntu.edu.sg)

(Corresponding author: Chen Lv)

### A. Decision Making for AVs

Many methods have been applied to the decision making of AVs in recent years. A probabilistic model is widely used to deal with the uncertainties of decision making [3], [4]. In [5], a Bayesian networks model is applied to decision making, that considers the uncertainties from environmental perception to decision making. However, it is difficult to deal with complex and dynamic decision missions. The Markov Decision Process (MDP) is another common probabilistic model for decision making. In [6], a robust decision-making approach is studied using partially observable MDP considering uncertain measurements, which can make AVs move safely and efficiently amid pedestrians. In [7], a stochastically verifiable decision-making framework is designed for AVs based on Probabilistic Timed Programs (PTPs) and Probabilistic Computational Tree Logic (PCTL). In addition, the Multiple Criteria Decision Making (MCDM) method and the Multiple Attribute Decision Making (MADM) method are effective to deal with complex urban environment and make reasonable decisions for AVs [8], [9].

With the development of machine learning algorithms, learning-based decision-making methods have been widely studied. In [10], a kernel-based extreme learning machine (KELM) modeling method is proposed for speed decision making of AVs. Since the machine learning algorithm is developed based on probabilistic inference rather than causal inference, as a result, it is difficult to understand and find the failure cause of the algorithm. In [11], a support vector machine (SVM) algorithm is applied to automatic decision making with Bayesian parameters optimization, which can deal with complex traffic scenarios. A novel decision-making system is built based on Deep Neural Networks (DNNs) in [12], which can adapt to real-life road conditions. However, training of the DNNs requires a large number of data samples. In [13], a stochastic MDP is used to model the interaction between an AV and the environment, and reinforcement learning (RL) is then applied to decision making based on the reward function of MDP. Compared with DNNs and other learning algorithms, the RL method does not need a large size of driving dataset, instead, it leverages a self-exploration mechanism that solves sequential decision-making problem via interacting with the environment [14]. A Q-learning (QL) decision-making method is proposed based on RL in [15]. However, the learning efficiency and generalization ability of the QL need to be further improved. To sum up, the learning-based decision-making methods, whose performances are limited by the quality of the dataset, still require further improvement.

In addition to the above learning-based methods, rule-based decision-making approaches, including the expert system and fuzzy logic, have been studied for a long time. In [16], a prediction and cost function-based expert system is proposed to deal with the decision-making problem of highway driving for AVs. However, the expert system developed is only able to deal with some simple scenarios. A decision-making approach using fuzzy logic is studied to address automatic overtaking in [17]. Similarly, it fails to handle various scenarios. In general, although the rule-based decision-making methods are easy to explain and understand, they would encounter difficulty with modelling and logic design, especially when the complexity of the system increases.

*B. Human-like Decision Making for Autonomous Driving*

It can be expected that human-driven vehicles and AVs will coexist on the roads in the coming decades. Driving safety will not be the only requirement for AVs, and how to interact with human-driven vehicles in complex traffic conditions is also vital and worthwhile studying. AVs are expected to behave similarly to human-driven vehicles. To this end, the behaviors and characteristics of human drivers should be considered in the automated driving design. If AVs' driving behaviors are considered as human-like, it would be easier for human drivers to interact with surrounding AVs and predict their behaviors, particularly with regard to multi-vehicle cooperation [18].

In recent years, some studies have introduced the human-like elements, typically represented by driving characteristics and driving styles, into the algorithm development for autonomous driving [19]-[21]. Specifically, in terms of the decision-making functionality of AVs, the human-like model is expected to generate reasonable driving behaviors similar to our human drivers. In [22], combining a deep autoencoder network and the XGBoost algorithm, a novel lane-change decision model is proposed, which can make human-like decisions for AVs. In [23], a multi-point turn decision making framework is designed based on the combination of the real human driving data and the vehicle dynamics for human-like autonomous driving. Learning from human driver's strategies for handling complex situations with potential risks, a human-like decision making algorithm is studied in [24]. To address the decision-making problem as well as the interactions with surrounding vehicles, a human-like behavior generation approach is proposed for AVs using the Markov decision process approach with hybrid potential maps [25]. It is able to ensure driving safety, generate efficient decisions as well as appropriate paths for AVs. Additionally, game theory is another effective way to formulate the human-like decision-making for AVs with social interactions. A game theoretic lane-change model is built in [26], which can provide a human-like manner during lane change process of AVs. In [27], the interactions between the host vehicle and surrounding ones are captured in a formulated game, which is supposed to make an optimal lane-change decision to overtake, merge and avoid collisions. Although the game theoretic approaches are capable of dealing with the decision-making for AVs under complex interactions, related studies considering different driving styles were rarely reported.

In fact, different drivers present individual preferences on driving safety, ride comfort and travel efficiency, indicating their different driving styles. For instance, in terms of the collision avoidance in emergency conditions, aggressive drivers may choose a fast steering response, while timid drivers may choose to operate the braking pedal. Additionally, different passengers may also request diverse driving styles. For instance, if pregnant women, elderly people as well as children are onboard, a more comfortable and safer riding experience is expected. In contrast, for daily commuters and those who are rushing to catch flights, travel efficiency takes a higher priority during their trips. Therefore, from the driving style perspective, the human-like decision-making of AVs is expected to provide personalized options for passengers with diverse requirements on driving safety, ride comfort and travel efficiency.

In this study, to further advance the human-like decision-making algorithms for AVs, typical driving styles are defined at first to reflect different driving characteristics during the modeling phase. Beyond this, the decision-making problem of AVs is formulated using game theory. Two noncooperative game approaches, i.e., the Nash equilibrium and Stackelberg games, are adopted to address decision-making and interactions for AVs. Finally, the developed human-like decision making algorithms for AVs are tested and validated via simulation in various scenarios.

*C. Contribution*

The contributions of this paper are summarized as follows: (1) The featured patterns and characteristics of three driving styles are identified and extracted from the real-world driving data. Based on the above findings, the different driving styles with their characteristics of social interactions are embedded into the models developed for decision making. (2) Two noncooperative game methods, i.e. the Nash equilibrium and Stackelberg game, are utilized for the design of the human-like decision making for AVs. It is found that both methods are able to generate reasonable and proper decisions for AVs, while the Stackelberg game is advantageous over the other one with respect to the resultant performance of the AV during decision making. (3) The high-accuracy motion prediction and collision -free path generation of AVs can be realized by using the developed algorithms based on the MPC with potential field.

*D. Paper Organization*

The rest of this paper is structured as follows. In Section II, the human-like decision-making framework is proposed. Then, the integrated driver-vehicle model is built in Section III. In Section IV, the decision-making module is established based on the noncooperative game theory. Next, the motion prediction and planning module is designed using MPC with the potential field method in Section V. Test results and analysis of the proposed approach are discussed in Section VI. Finally, Section VII concludes the paper.

## II. Human-like Decision-making Framework

### A. System Framework for Human-like Decision Making

The proposed human-like decision making framework for AVs in the lane-change scenario is illustrated in Fig. 1. It mainly consists of three modules, i.e., modeling, decision making, and motion planning. In the modeling module, three different driving styles, namely, the aggressive, normal and conservative styles, are defined first. Then, a driver model is integrated to the vehicle-road model with consideration of the pre-defined driving styles to generate human-like features in autonomous driving. Based on this human-like model built, the cost functions for decision making are constructed, in which three indexes of the driving performance, including driving safety, ride comfort and travel efficiency, are adopted. Furthermore, two noncooperative game theoretic approaches are utilized to address the lane-change decision making issue of AVs. Afterward, the decision-making commands are sent to the motion planning module. Combined with the potential model method, the MPC is applied for motion prediction and planning, which aims to execute the decision made and plan collision-avoidance paths for the AV. Additionally, the motion states and positions of surrounding vehicles should be provided to the decision making and motion planning modules of the ego car. Through the above manners, the human-like features are embedded and reflected in both the modeling and decision-making of AVs.

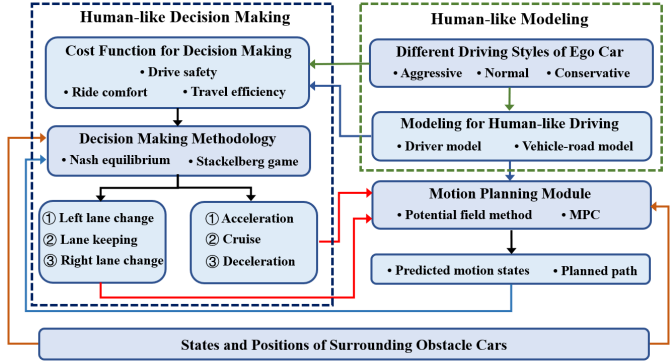

Fig. 1. Human-like decision making framework for AVs.

### B. Human-like Decision Making Modeling

In the human-like modeling module, three driving styles, i.e., aggressive, normal and conservative, are considered for AVs. For human drivers, driving styles are associated with both human factors and environment conditions, including human physical limitations, habits, personalities, ages, road conditions, weathers, etc. [28], [29]. Typically, aggressive drivers care more care about the vehicle dynamic performance and travel efficiency. Usually, they would frequently adjust the steering wheel and operate the throttle and brake pedals. In contrast, conservative drivers give the priority to safety and ride comfort. Therefore, they execute control actions much more carefully. While for normal drivers, who are positioned between the above two styles, are likely to make a trade-off, coordinating the travel efficiency, ride comfort and safety [30], [31].

### C. Human-like Decision Making Methodology

In the human-like decision making module, two noncooperative game theoretic approaches, i.e., the Nash equilibrium and Stackelberg game, are used to simulate the human-like decision-making and interaction behaviors for AVs. In the decision-making problem with the Nash equilibrium, all AVs are seen as equal and independent players. When solving the formulated problem, all AVs aim to minimize their own cost functions without considering the others' payoffs. However, when solving the decision-making problem using the Stackelberg game, the ego car is considered as a lead player, and surrounding cars are regarded as followers. Thus, the decision making of the ego car considers the behaviors of all surrounding cars.

## III. Modeling for Human-like Driving

In this section, the characteristics of different driving styles are analyzed and extracted, which are reflected in the driver model. Then, the combination of the driver model and vehicle-road model yields an integrated model for human-like driving, which is applied to the decision making and motion planning of AVs.

### A. Characteristic Analysis of Different Driving Styles

To analyze and extract the characteristics of human driving styles, a real-world driving dataset, i.e. the NGSIM, is used. The NGSIM vehicle data are collected from different regions in different time slots, which reflects various traffic scenarios [32]. In this study, two groups of driving data from NGSIM, i.e., the I-80 and US-101 freeway sub-datasets, are utilized for analysis.

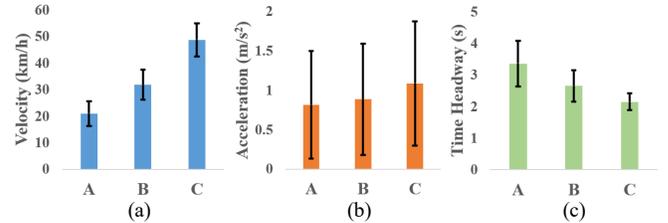

Fig. 2. Characteristic analysis of different driving styles with the NGSIM data: (a) velocity; (b) acceleration; (c) time headway. The three driving styles are denoted by, A: conservative, B: normal, and C: aggressive, respectively.

As mentioned above, three indexes of driving performance, including the velocity, acceleration and time headway, are analyzed. The vehicle velocity is adopted to represent the travel efficiency, and the vehicle acceleration is associated with the ride comfort, while the time headway reflects the driving safety. The mean and standard deviation (STD) values of the performance indexes are analyzed and illustrated in Fig. 2. It shows that aggressive human drivers tend to have a higher velocity, a larger acceleration, and a smaller time headway, indicating that they put more weights on the travel efficiency rather than safety and comfort. However, it comes with an opposite conclusion for the conservative drivers regarding the above pattern. In addition, the pattern of the three indexes for normal divers are positioned between that of the aggressive and conservative drivers. The above analysis and features extracted regarding the human driving styles can be utilized in the cost function design of human-like decision making for AVs.

*B. Driver Model*

Fig. 3 shows a single-point preview driver model, that is widely used in vehicle control [33]. In this paper, the model is applied to the motion prediction of AVs. In Fig. 3, point $E$ is the current position of the driver. $\varphi$ is the yaw angle of the vehicle. $M$ is the predicted point along the current moving direction of the vehicle in the future, which is predicted by the driver's brain according to the vehicle states and position. $P$ is the preview point created by the driver's eyes, and consecutive preview points make up the planned path. The driver aims to minimize the distance between the predicted point and the preview point. Finally, the purpose is realized by controlling the steering angle of the front wheel. The aforementioned content describes the basic working principle of the driver model.

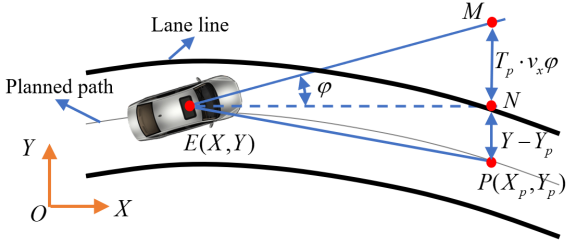

Fig. 3. Driver model.

Considering the driver's driving characteristics, the driver model is expressed as follows [34].

$$\ddot{\delta}_f = -\frac{1}{aT_d}\dot{\delta}_f - \frac{1}{aT_d^2}\delta_f + \frac{K_s G_s}{aT_d^2}[Y_p - (Y + T_p \cdot v_x \varphi)] \quad (1)$$

where $\delta_f$ is the steering angle of the front wheel in the bicycle model, and $K_s$ is a proportional coefficient associated with $\delta_f$ and the steering wheel angle, representing the transmission ratio of steering system. $a$ is related to the damping rate of the model, $T_d$ and $T_p$ are the driver's physical delay time and predicted time, $v_x$ is the longitudinal velocity of the vehicle, $Y$ and $Y_p$ are the lateral coordinate position of the vehicle and preview point, and $G_s$ is the steering proportional gain. The driver's driving style or characteristic is reflected by $T_d$, $T_p$ and $G_s$. For instance, conservative drivers require more time for mental signal processing and muscular activation than aggressive drivers. Therefore, conservative drivers have larger $T_d$. However, aggressive drivers have opposite reactions [35].

Referring to [36]-[38], the parameters that reflect different driving styles are selected in Table I.

TABLE I
PARAMETERS OF DIFFERENT DRIVING STYLES

| Parameters | Aggressive | Normal | Conservative |
|---|---|---|---|
| $T_d$ | 0.14 | 0.18 | 0.24 |
| $T_p$ | 1.02 | 0.94 | 0.83 |
| $G_s$ | 0.84 | 0.75 | 0.62 |
| $a$ | 0.24 | 0.23 | 0.22 |

*C. Vehicle-Road Model*

To reduce the complexity of the controller design, the four-wheel vehicle model is simplified as a two-wheel bicycle model [39], which is illustrated in Fig. 4.

Assuming that the steering angle of the front wheel $\delta_f$ has a small value, it yields $\sin \delta_f \approx 0$. According to Fig. 5, the simplified vehicle dynamic model is derived as follows [40].

$$\begin{cases} \dot{v}_x = v_y r + F_{xf} \cos \delta_f /m + F_{xr}/m \\ \dot{v}_y = -v_x r + F_{yf} \cos \delta_f /m + F_{yr}/m \\ \dot{r} = l_f F_{yf} \cos \delta_f /I_z - l_r F_{yr}/I_z \end{cases} \quad (2)$$

where

$$F_{xf} \cos \delta_f /m + F_{xr}/m = a_x \quad (3)$$

$$F_{yf} = -k_f \alpha_f, \quad F_{yr} = -k_r \alpha_r \quad (4)$$

$$\alpha_f = -\delta_f + (v_y + l_f r)/v_x, \quad \alpha_r = (v_y - l_r r)/v_x \quad (5)$$

$v_y$ is the lateral velocity of the vehicle, and $r$ is the yaw rate. $F_{xi}(i = f, r)$ and $F_{yi}(i = f, r)$ denote the longitudinal and lateral tire forces of the front and rear wheels. $l_f$ and $l_r$ are the front and rear wheel bases, respectively. $m$ is the vehicle mass, and $I_z$ is the yaw moment of inertia. $k_f$ and $k_r$ are the cornering stiffness of the front and rear tires, respectively. $\alpha_f$ and $\alpha_r$ are the slip angles of the front and rear tires, respectively.

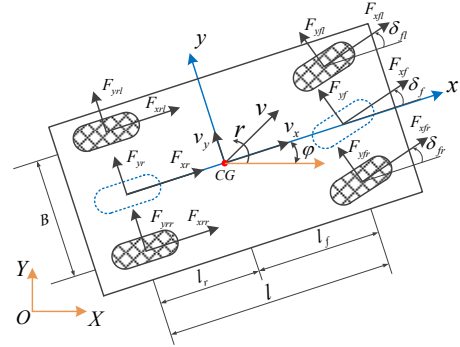

Fig. 4. Bicycle model.

The vehicle parameters of the bicycle model are listed in Table II.

TABLE II
PARAMETERS OF THE BICYCLE MODEL

| Parameter | Value | Parameter | Value |
|---|---|---|---|
| $m$/(kg) | 1300 | $I_z$/(kg·m$^2$) | 2500 |
| $l_f$/(m) | 1.25 | $l_r$/(m) | 1.32 |
| $k_f$/(N/rad) | 35000 | $k_r$/(N/rad) | 38000 |

In addition, the vehicle kinematic model in global coordinates is expressed as

$$\begin{cases} \dot{X} = v_x \cos \varphi - v_y \sin \varphi \\ \dot{Y} = v_x \sin \varphi + v_y \cos \varphi \\ \dot{\varphi} = r \end{cases} \quad (6)$$

Combining the driver model and the vehicle-road model yields the integrated model, which is expressed as

$$\dot{x}(t) = f[x(t), u(t)] \quad (7)$$

$$f = \begin{bmatrix} v_y r + a_x \\ -v_x r + F_{yf} \cos \delta_f /m + F_{yr}/m \\ l_f F_{yf} \cos \delta_f /I_z - l_r F_{yr}/I_z \\ r \\ v_x \cos \varphi - v_y \sin \varphi \\ v_x \sin \varphi + v_y \cos \varphi \\ \dot{\delta}_f \\ -\frac{1}{aT_d^2}\delta_f - \frac{1}{aT_p}\dot{\delta}_f + \frac{R_g G_h}{aT_d^2}[Y_p - (Y - T_p \cdot v_x \varphi)] \end{bmatrix} \quad (8)$$

where the state vector $x = [v_x, v_y, r, \varphi, X, Y, \delta_f, \dot{\delta}_f]^T$, and the control vector $u = Y_p$.

## IV. DECISION MAKING BASED ON NONCOOPERATIVE GAME THEORY

In this section, modeling for lane-change decision making is conducted considering different driving styles. Then, Nash equilibrium and Stackelberg game theory are applied to the noncooperative decision making. In the modeling process, the lane-change behaviors of obstacle vehicles are not included, thus only acceleration and deceleration behaviors are considered for obstacle vehicles.

### A. Cost Function for Lane-change Decision Making

Fig. 5 shows a common lane-change scenario on a three-lane highway. Ego car (EC) is an AV moving on the middle lane, and lead car 2 (LC2) moves in front of EC with a slow velocity. EC has to decide to decelerate and follow LC2 or change lanes, then, change to the left lane or the right lane, which is a decision-making problem.

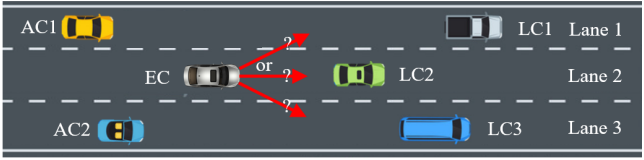

Fig. 5. Lane-change decision making.

Three performance indexes are taken into consideration in the modeling process of decision making, i.e., driving safety, ride comfort and travel efficiency. Therefore, the cost function of decision making for EC is the integration of these three components, which is expressed as

$$J^{EC} = w_{ds}^{EC} J_{ds}^{EC} + w_{rc}^{EC} J_{rc}^{EC} + w_{pe}^{EC} J_{pe}^{EC} \quad (9)$$

where $J_{ds}^{EC}$, $J_{rc}^{EC}$ and $J_{pe}^{EC}$ denote the costs of driving safety, ride comfort and travel efficiency, respectively. $w_{ds}^{EC}$, $w_{rc}^{EC}$ and $w_{pe}^{EC}$ are the weighting coefficients.

The driving safety cost of EC is made up of two parts, i.e., the longitudinal driving safety cost and the lateral driving safety cost, which are related to the lead car (LC) and adjacent car (AC), respectively. Furthermore, driving safety cost $J_{ds}^{EV}$ is defined by

$$J_{ds}^{EC} = (\sigma^2 - 1)^2 J_{ds-log}^{EC} + \sigma^2 J_{ds-lat}^{EC} \quad (10)$$

where $J_{ds-log}^{EC}$ and $J_{ds-lat}^{EC}$ denote the longitudinal driving safety cost and the lateral driving safety cost, respectively. $\sigma$ is the lane-change behavior of EC, $\sigma \in \{-1,0,1\}:=\{$change left, no lane change, change right$\}$.

The longitudinal driving safety cost $J_{ds-log}^{EC}$ is a function of the longitudinal gap and relative velocity with respective to LC, which is defined by

$$J_{ds-log}^{EC} = \kappa_{v-log}^{EC} \lambda_v^{EC} (\Delta v_{x,v}^{EC})^2 + \kappa_{s-log}^{EC}/[(\Delta s_{x,v}^{EC})^2 + \varepsilon] \quad (11a)$$
$$\Delta v_{x,v}^{EC} = v_{x,v}^{LC} - v_{x,v}^{EC} \quad (11b)$$
$$\Delta s_{x,v}^{EC} = \sqrt{(X_v^{LC} - X_v^{EC})^2 + (Y_v^{LC} - Y_v^{EC})^2} - l_v \quad (11c)$$
$$\lambda_v^{EC} = \begin{cases} 1, & \Delta v_{x,v}^{EC} < 0 \\ 0, & \Delta v_{x,v}^{EC} \geq 0 \end{cases} \quad (12d)$$

where $v_{x,v}^{LC}$ and $v_{x,v}^{EC}$ denote the longitudinal velocities of LC and EC, respectively. $(X_v^{LC}, Y_v^{LC})$ and $(X_v^{EC}, Y_v^{EC})$ are the positions of LC and EC, respectively. $\kappa_{v-log}^{EC}$ and $\kappa_{s-log}^{EC}$ are the weighting coefficients. $\varepsilon$ is a very small value to avoid zero denominator in the calculation. $l_v$ is a safety coefficient considering the length of the car. $v$ denotes the lane number, $v \in \{1,2,3\}:=\{$left lane, middle lane, right lane$\}$.

The lateral driving safety cost $J_{ds-lat}^{EC}$ is associated with the longitudinal gap and relative velocity with respective to AC, which is expressed as

$$J_{ds-lat}^{EC} = \kappa_{v-lat}^{EC} \lambda_{v+\sigma}^{EC} (\Delta v_{x,v+\sigma}^{EC})^2 + \kappa_{s-lat}^{EC}/[(\Delta s_{x,v+\sigma}^{EC})^2 + \varepsilon] \quad (12a)$$
$$\Delta v_{x,v+\sigma}^{EC} = v_{x,\delta}^{EC} - v_{x,v+\sigma}^{AC} \quad (12b)$$
$$\Delta s_{x,v+\sigma}^{EC} = \sqrt{(X_v^{EC} - X_{v+\sigma}^{AC})^2 + (Y_v^{EC} - Y_{v+\sigma}^{AC})^2} - l_v \quad (12c)$$
$$\lambda_{v+\sigma}^{EC} = \begin{cases} 1, & \Delta v_{x,v+\sigma}^{EC} < 0 \\ 0, & \Delta v_{x,v+\sigma}^{EC} \geq 0 \end{cases} \quad (12d)$$

where $v_{x,v+\sigma}^{AC}$ is the longitudinal velocity of AC, $(X_{\delta+\alpha}^{AC}, Y_{\delta+\alpha}^{AC})$ is the position of AC, $\kappa_{v-lat}^{EC}$ and $\kappa_{s-lat}^{EC}$ are the weighting coefficients.

The ride comfort cost $J_{rc}^{EC}$ is directly related to the longitudinal acceleration and lateral acceleration, which is defined by

$$J_{rc}^{EC} = \kappa_{a_x}^{EC}(a_{x,v}^{EC})^2 + \sigma^2 \kappa_{a_y}^{EC}(a_{y,v}^{EC})^2 \quad (13)$$

where $a_{x,v}^{EC}$ and $a_{y,v}^{EC}$ are the longitudinal acceleration and lateral acceleration of EC. $\kappa_{a_x}^{EC}$ and $\kappa_{a_y}^{EC}$ are the weighting coefficients.

The travel efficiency cost $J_{pe}^{EC}$ is associated with the longitudinal velocity of EC, which is expressed as

$$J_{pe}^{EC} = (v_{x,v}^{EC} - \bar{v}_{x,v}^{EC})^2, \quad \bar{v}_{x,v}^{EC} = \min(v_{x,v}^{max}, v_{x,v}^{LC}) \quad (14)$$

where $v_{x,v}^{max}$ is the velocity limit on the lane $v$.

Furthermore, the cost function for AC has a similar expression, which will not be introduced repeatedly. Two differences are noted. First, since it is assumed that AC does not change lanes and only assumes acceleration or deceleration behavior, the lateral driving safety cost of AC is equal to the lateral driving safety cost of EC. The ride comfort cost of AC only comes from longitudinal acceleration.

As mentioned in Section II, three different driving styles, i.e., aggressive, normal and conservative, are defined to describe the social behaviors of AV. In the modeling process of decision making, the differences between the different driving styles are reflected to different settings of weighting coefficients, i.e., $w_{ds}^{EC}$,

$w_{rc}^{EC}$ and $w_{pe}^{EC}$, which are related to driving safety, ride comfort and travel efficiency. Referring to [30], [31], the weighting coefficients of the three driving styles are shown in Table III.

TABLE III
WEIGHTING COEFFICIENTS OF THE DIFFERENT DRIVING STYLES

| Weighting Coefficients | Aggressive | Normal | Conservative |
|---|---|---|---|
| $w_{ds}^{EC}$ | 10% | 50% | 70% |
| $w_{rc}^{EC}$ | 10% | 30% | 20% |
| $w_{pe}^{EC}$ | 80% | 20% | 10% |

### B. Noncooperative Decision Making Based on Nash Equilibrium

In common lane-change scenarios, the game problem occurs between EC and AC, which is a 2-player game problem. The noncooperative decision making based on Nash equilibrium for two players can be expressed as

$$(a_{x,v}^{EC*}, \sigma^*) = \arg \min_{a_{x,v}^{EC}, \sigma} J^{EC}(a_{x,v}^{EC}, \sigma, a_{x,v+\sigma}^{AC}) \quad (15a)$$

$$a_{x,v+\sigma}^{AC*} = \arg \min_{a_{x,v+\sigma}^{AC}} J^{AC}(a_{x,v}^{EC}, \sigma, a_{x,v+\sigma}^{AC}) \quad (15b)$$

s.t.

$\sigma \in \{-1,0,1\}, a_{x,v}^{EC} \in [a_{x,v}^{\min}, a_{x,v}^{\max}], a_{x,v+\sigma}^{AC} \in [a_{x,v+\sigma}^{\min}, a_{x,v+\sigma}^{\max}],$
$v_{x,v}^{EC} \in [v_{x,v}^{\min}, v_{x,v}^{\max}], v_{x,v+\sigma}^{AC} \in [v_{x,v+\sigma}^{\min}, v_{x,v+\sigma}^{\max}].$

where $a_{x,v}^{EC*}$ and $a_{x,v+\sigma}^{AC*}$ are the optimal longitudinal accelerations of EC and AC, $\sigma^*$ is the optimal lane-change behavior of EC, $a_{x,i}^{min}$ and $a_{x,i}^{max}$ are the minimum and maximum boundaries of longitudinal acceleration, $v_{x,i}^{min}$ and $v_{x,i}^{max}$ are the minimum and maximum constraints of longitudinal velocity.

If there are two ACs on the left and right lanes respectively as Fig. 5 shows, Eq. (15) should be rewritten as

$$(a_{x,v}^{EC*}, \sigma^*) = \arg \min_{(a_{x,v}^{EC}, \sigma) \in A} [J_1, J_2] \quad (16a)$$

$$A = \{(a_{x,v,1}^{EC}, \sigma_1), (a_{x,v,2}^{EC}, \sigma_2)\} \quad (16b)$$

$$J_1 = \min_{a_{x,v,1}^{EC}, \sigma_1} J^{EC}(a_{x,v,1}^{EC}, \sigma_1, a_{x,v+\sigma_1}^{AC1}) \quad (16c)$$

$$J_2 = \min_{a_{x,v,2}^{EC}, \sigma_2} J^{EC}(a_{x,v,2}^{EC}, \sigma_2, a_{x,v+\sigma_2}^{AC2}) \quad (16d)$$

$$a_{x,v+\sigma_1}^{AC1*} = \arg \min_{a_{x,v+\sigma_1}^{AC1}} J^{AC1}(a_{x,v,1}^{EC}, \sigma_1, a_{x,v+\sigma_1}^{AC1}) \quad (16e)$$

$$a_{x,v+\sigma_2}^{AC2*} = \arg \min_{a_{x,v+\sigma_2}^{AC2}} J^{AC2}(a_{x,v,2}^{EC}, \sigma_2, a_{x,v+\sigma_2}^{AC2}) \quad (16f)$$

s.t.

$a_{x,v,1}^{EC}, a_{x,v,2}^{EC} \in [a_{x,v}^{\min}, a_{x,v}^{\max}], v_{x,v,1}^{EC}, v_{x,v,2}^{EC} \in [v_{x,v}^{\min}, v_{x,v}^{\max}],$
$a_{x,v+\sigma_1}^{AC1} \in [a_{x,v+\sigma_1}^{\min}, a_{x,v+\sigma_1}^{\max}], a_{x,v+\sigma_2}^{AC2} \in [a_{x,v+\sigma_2}^{\min}, a_{x,v+\sigma_2}^{\max}],$
$v_{x,v+\sigma_1}^{AC1} \in [v_{x,v+\sigma_1}^{\min}, v_{x,v+\sigma_1}^{\max}], v_{x,v+\sigma_2}^{AC2} \in [v_{x,v+\sigma_2}^{\min}, v_{x,v+\sigma_2}^{\max}],$
$\sigma_1, \sigma_2 \in \{-1,0,1\}.$

### C. Noncooperative Decision Making Based on Stackelberg Equilibrium

In addition to Nash equilibrium, Stackelberg game is another noncooperative game approach [41], [42]. In Nash equilibrium, EC and AC are two equal and independent players. In the solving process, both EC and AC aim to minimize their own cost functions of decision making without considering the opponent's decision-making results. However, in Stackelberg equilibrium, EC is a leader player and AC is a follower player, the behavior of AC will affect the decision making of EC. Applying Stackelberg equilibrium into lane-change decision making, it yields

$$(a_{x,v}^{EC*}, \sigma^*) = \arg \min_{a_{x,v}^{EC}, \sigma} ( \max_{a_{x,v+\sigma}^{AC} \in \gamma^2(a_{x,v}^{EC}, \sigma)} J^{EC}(a_{x,v}^{EC}, \sigma, a_{x,v+\sigma}^{AC})) \quad (17a)$$

$$\gamma^2(a_{x,v}^{EC}, \sigma) = \{\zeta \in \Phi^2 : J^{AC}(a_{x,v}^{EC}, \sigma, \zeta) \leq J^{AC}(a_{x,v}^{EC}, \sigma, a_{x,v+\sigma}^{AC}), \forall a_{x,v+\sigma}^{AC} \in \Phi^2\} \quad (17b)$$

s.t.

$\sigma \in \{-1,0,1\}, a_{x,v}^{EC} \in [a_{x,v}^{\min}, a_{x,v}^{\max}], a_{x,v+\sigma}^{AC} \in [a_{x,v+\sigma}^{\min}, a_{x,v+\sigma}^{\max}],$
$v_{x,v}^{EC} \in [v_{x,v}^{\min}, v_{x,v}^{\max}], v_{x,v+\sigma}^{AC} \in [v_{x,v+\sigma}^{\min}, v_{x,v+\sigma}^{\max}].$

Similarly, if there are two ACs on the left and right lanes respectively as Fig. 5 shows, Eq. (17) should be rewritten as

$$(a_{x,v}^{EC*}, \sigma^*) = \arg \min_{(a_{x,v}^{EC}, \sigma) \in A} [J^{EC}(a_{x,v,1}^{EC}, \sigma_1, a_{x,v+\sigma_1}^{AC1}), J^{EC}(a_{x,v,2}^{EC}, \sigma_2, a_{x,v+\sigma_2}^{AC2})] \quad (18a)$$

$$A = \{(a_{x,v,1}^{EC}, \sigma_1), (a_{x,v,2}^{EC}, \sigma_2)\} \quad (18b)$$

$$(a_{x,v,1}^{EC}, \sigma_1) = \arg \min_{a_{x,v,1}^{EC}, \sigma_1} ( \max_{a_{x,v+\sigma_1}^{AC1} \in \gamma_1^2(a_{x,v,1}^{EC}, \sigma_1)} J^{EC}(a_{x,v,1}^{EC}, \sigma_1, a_{x,v+\sigma_1}^{AC1})) \quad (18c)$$

$$\gamma_1^2(a_{x,v,1}^{EC}, \sigma_1) = \{\zeta_1 \in \Phi_1^2 : J^{AC1}(a_{x,v,1}^{EC}, \sigma_1, \zeta_1) \leq J^{AC1}(a_{x,v,1}^{EC}, \sigma_1, a_{x,v+\sigma_1}^{AC1}), \forall a_{x,v+\sigma_1}^{AC1} \in \Phi_1^2\} \quad (18d)$$

$$(a_{x,v,2}^{EC}, \sigma_2) = \arg \min_{a_{x,v,2}^{EC}, \sigma_2} ( \max_{a_{x,v+\sigma_2}^{AC2} \in \gamma_2^2(a_{x,v,2}^{EC}, \sigma_2)} J^{EC}(a_{x,v,2}^{EC}, \sigma_2, a_{x,v+\sigma_2}^{AC2})) \quad (18e)$$

$$\gamma_2^2(a_{x,v,2}^{EC}, \sigma_2) = \{\zeta_2 \in \Phi_2^2 : J^{AC2}(a_{x,v,2}^{EC}, \sigma_2, \zeta_2) \leq J^{AC2}(a_{x,v,2}^{EC}, \sigma_2, a_{x,v+\sigma_2}^{AC2}), \forall a_{x,v+\sigma_2}^{AC2} \in \Phi_2^2\} \quad (18f)$$

s.t.

$a_{x,v,1}^{EC}, a_{x,v,2}^{EC} \in [a_{x,v}^{\min}, a_{x,v}^{\max}], v_{x,v,1}^{EC}, v_{x,v,2}^{EC} \in [v_{x,v}^{\min}, v_{x,v}^{\max}],$
$a_{x,v+\sigma_1}^{AC1} \in [a_{x,v+\sigma_1}^{\min}, a_{x,v+\sigma_1}^{\max}], a_{x,v+\sigma_2}^{AC2} \in [a_{x,v+\sigma_2}^{\min}, a_{x,v+\sigma_2}^{\max}],$
$v_{x,v+\sigma_1}^{AC1} \in [v_{x,v+\sigma_1}^{\min}, v_{x,v+\sigma_1}^{\max}], v_{x,v+\sigma_2}^{AC2} \in [v_{x,v+\sigma_2}^{\min}, v_{x,v+\sigma_2}^{\max}],$
$\sigma_1, \sigma_2 \in \{-1,0,1\}.$

## V. MOTION PREDICTION AND PLANNING BASED ON THE POTENTIAL FIELD MODEL METHOD AND MPC

The potential field method is able to deal with the dynamic modeling of obstacles and roads, which is an effective method for motion planning. In this section, the potential field method is combined with MPC to predict the motion states and collision-avoidance path for the decision-making framework.

### A. Potential Field Method for Collision Avoidance

The integrated potential field function of obstacle cars (OCs) and roads is expressed as

$$\Gamma(X, Y) = \sum_{i=1}^{n_1} \Gamma_i^{oc}(X, Y) + \sum_{j=1}^{n_2} \Gamma_j^r(X, Y) \quad (19)$$

where $n_1$ and $n_2$ are the numbers of OCs and lane lines, respectively. OCs include LVs and ACs.

$\Gamma^{oc}(X, Y)$ is the potential field function of the OC at the position $(X, Y)$ in the global coordinates, it is defined as [43]

$$\Gamma^{oc}(X, Y) = a^{oc} \cdot \exp(\vartheta) \quad (20)$$

where

$$\vartheta = -\left\{\frac{\hat{X}^2}{2\rho_X^2} + \frac{\hat{Y}^2}{2\rho_Y^2}\right\}^b + cv_x^{oc}\gamma, \gamma = k^{oc}\frac{\hat{X}^2}{2\rho_X^2} / \sqrt{\frac{\hat{X}^2}{2\rho_X^2} + \frac{\hat{Y}^2}{2\rho_Y^2}},$$

$$k^{oc} = \begin{cases} -1, & \hat{X} < 0 \\ 1, & \hat{X} \geq 0 \end{cases}, \begin{bmatrix} \hat{X} \\ \hat{Y} \end{bmatrix} = \begin{bmatrix} \cos \varphi^{oc} & \sin \varphi^{oc} \\ -\sin \varphi^{oc} & \cos \varphi^{oc} \end{bmatrix} \begin{bmatrix} X - X^{oc} \\ Y - Y^{oc} \end{bmatrix}.$$

$(X^{oc}, Y^{oc})$ is the CoG (center of gravity) position of OC. $a^{oc}$ is the maximum potential field value of OC. $\rho_X$ and $\rho_Y$ are the convergence coefficients. $\varphi^{oc}$ and $v_x^{oc}$ are the heading angle and longitudinal velocity of OC. $b$ is the shape coefficient.

$\Gamma^r(X, Y)$ is the potential field function of the road, which is defined by

$$\Gamma^r(X, Y) = a^r \cdot \exp(-d + d^\dagger + 0.5W) \tag{21}$$

where $a^r$ is the maximum potential field value of the road, $d$ is the minimum distance from the position $(X, Y)$ to the lane line, $d^\dagger$ is the safety threshold value and $W$ is the width of OC.

*B. Motion Prediction Based on MPC*

To conduct motion prediction using MPC, the integrated model, i.e., Eq. (8) is transformed into a time-varying linear system.

$$\dot{x}(t) = A_t x(t) + B_t u(t) \tag{22}$$

where the time-varying coefficient matrices are given by

$$A_t = \left.\frac{\partial f}{\partial x}\right|_{x_t, u_t}, \quad B_t = \left.\frac{\partial f}{\partial u}\right|_{x_t, u_t} \tag{23}$$

Furthermore, Eq. (22) can be discretized as

$$\begin{cases} x(k+1) = A_k x(k) + B_k u(k) \\ u(k) = u(k-1) + \Delta u(k) \end{cases} \tag{24}$$

where $A_k = e^{A_t \Delta T}$, $B_k = \int_0^{\Delta T} e^{A_t \tau} B_t d\tau$, $\Delta T$ is the sampling time, $x(k) = [v_{x,v}^{EC}(k), v_{y,v}^{EC}(k), r_v^{EC}(k), \varphi_v^{EC}(k), X_v^{EC}(k), Y_v^{EC}(k), \delta_{f,v}^{EC}(k), \dot{\delta}_{f,v}^{EC}(k)]^T$, $u(k) = Y_{p,v}^{EC}$, and $\Delta u(k) = \Delta Y_{p,v}^{EC}$.

In addition, the output vector is defined by $y(k) = [y_1(k), y_2(k), y_3(k)]^T$, where $y_1(k)$ is the potential field value associated with the predicted position of EC, i.e., $y_1(k) = \Gamma(X_v^{EC}(k), Y_v^{EC}(k))$, and $y_2(k)$ and $y_3(k)$ are the lateral distance error and yaw angle error between the predicted position of EC and the center line of the lane $v$, respectively.

Moreover, the predictive horizon $N_p$ and the control horizon $N_c$ are defined, $N_p \geq N_c$. At the time step $k$, the sate sequence, output sequence and control sequence are given by

$$x(k+1|k), x(k+2|k), \cdots, x(k+N_p|k) \tag{25a}$$
$$y(k+1|k), y(k+2|k), \cdots, y(k+N_p|k) \tag{25b}$$
$$\Delta u(k|k), \Delta u(k+1|k), \cdots, \Delta u(k+N_c-1|k) \tag{25c}$$

The cost function for motion prediction of EC is defined by

$$\Theta(k) = \sum_{i=1}^{N_p} \|y(k+i|k)\|_Q^2 + \sum_{j=0}^{N_c-1} \|\Delta u(k+j|k)\|_R^2 \tag{26}$$

where $Q$ is the output weighting matrix, and $R$ is the control variation weighting matrix.

Finally, the motion prediction and planning problem of EC can be expressed as

$$\Delta \mathbf{u}(k) = \arg \min_{\Delta \mathbf{u}(k)} \Theta(k) \tag{27}$$

s.t.

$$x(k+i|k) = A_k x(k+i-1|k) + B_k u(k+i-1|k)$$
$$y(k+i-1|k) = [y_1(k+i-1|k), y_2(k+i-1|k),$$
$$y_3(k+i-1|k)]^T, (i = 1, 2, \cdots, N_p),$$
$$u(k+i|k) = u(k+i-1|k) + \Delta u(k+i|k),$$
$$u(k+i|k) \in [u_{\min}, u_{\max}],$$
$$\Delta u(k+i|k) \in [\Delta u_{\min}, \Delta u_{\max}], (i = 0, 1, \cdots, N_c - 1).$$

where $u_{min}$ and $u_{max}$ are the minimum and maximum constraints of $u$, $\Delta u_{min}$ and $\Delta u_{max}$ are the minimum and maximum constraints of $\Delta u$. After solving the optimization problem, the sequence of optimal input increments is obtained.

$$\Delta \mathbf{u}^*(k) = [\Delta u^*(k|k), \Delta u^*(k+1|k), \cdots, \Delta u^*(k+N_c-1|k)] \tag{28}$$

Then, the control output at the time step $k$ can be calculated.

$$u(k|k) = u(k-1|k-1) + \Delta u^*(k|k) \tag{29}$$

Furthermore, the state vector is updated according to Eq. (22). In another words, the motion prediction of EC is realized. At the next time step $k+1$, a new optimization is solved over a shifted prediction horizon with the updated state $x(k+1|k+1)$.

## VI. Testing Results and Performance Evaluation

In this section, two testing scenarios are designed to evaluate the feasibility and effectiveness of the proposed human-like decision making framework for AVs. All the driving scenarios are built and tested on the MATLAB-Simulink platform.

*A. Scenario A*

As Fig. 6 shows, Scenario A is a merging maneuver on a highway. In this scenario, EC has to change lanes due to the impending ending of the current lane. However, there exits an AC on the left lane. The decision making of EC must consider the reaction behavior of AC. In addition, the different driving styles of EC have significant effects on decision making. At the initial moment, the longitudinal velocities of EC and AC are 20 m/s and 15 m/s, and the initial gap between EC and AC is 2 m (EC is in front of AC). The testing results are demonstrated in Figs. 7-9.

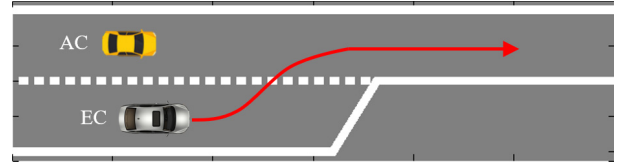

Fig. 6. Testing Scenario A.

It can be seen from Fig. 7, that the different driving styles and different strategies would lead to far different results regarding decision-making and motion planning. Also reflected by the numerical analysis in Table IV, the aggressive style has the shortest time for decision-making of merging, ensuring the travel efficiency, while the vehicle with conservative mode decelerate in the longitudinal direction, resulting in a longer travel time while guaranteeing the driving safety. It can be found from Table V that, in the three driving styles, the aggressive mode has the largest cost value of driving safety and the smallest cost value of travel efficiency, indicating that the aggressive mode cares more about travel efficiency than driving safety. However, the driving safety dominates the weighting in the conservative mode. The normal mode is found to make a balance among the travel efficiency, ride comfort and driving safety. Moreover, from the testing results of Figs. 8 and 9, similar conclusions can also be derived. If EC is aggressive, it would choose a sudden large acceleration to increase the gap immediately. Then, a lane-change behavior would be conducted. If the driving style of EC is normal, instead of speeding up with a large acceleration, then the EC would accelerate progressively and increase the gap gradually. and a lane-change maneuver would be carried out only when the gap is large enough. In contrast, the decision making pattern for the conservative mode is opposite to that of the aggressive one.

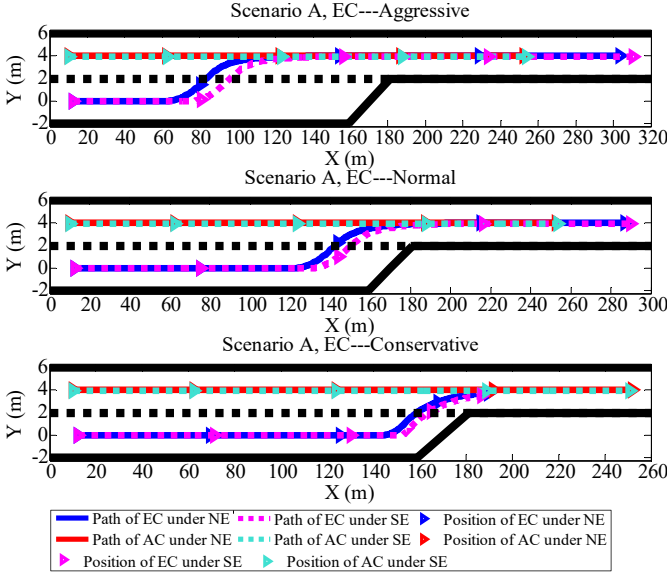

Fig. 7. Decision-making results in Scenario A.

Additionally, it can be found from the testing results in Figs. 7-9, the two different noncooperative game strategies, i.e., the Nash Equilibrium (NE) and Stackelberg Equilibrium (SE), would lead to different decision-making results. From Table IV, it is analyzed that the time for decision-making during merging of SE is longer than that of NE, which brings a larger merging velocity as well as a larger gap for EC. Moreover, as listed in Table V, the performance indicators under SE are much better than those of NE. In the normal driving style, the cost values of the driving safety, ride comfort and travel efficiency in SE are reduced by 22%, 20% and 22%, respectively. Specifically, in NE, the EC and AC are two equal but independent players. Both EC and AC aim to minimize their own cost functions during decision making. However, as the EC is the lead player in SE and AC is a follower player, the behavior of AC is fed back to the decision making of EC. As a result, EC is able to adjust its decision made and improve the driving performance based on the behaviors of surrounding vehicles accordingly.

TABLE IV
TESTING RESULTS OF DECISION MAKING IN SCENARIO A

| Testing Parameters | Aggressive | | Normal | | Conservative | |
|---|---|---|---|---|---|---|
| | NE | SE | NE | SE | NE | SE |
| $t_c$ / s | 2.24 | 2.72 | 5.04 | 5.36 | 7.00 | 7.18 |
| $\Delta s_{t_c}$ / m | 13.77 | 16.46 | 18.06 | 18.86 | -1.37 | -1.99 |
| $v_{x,t_c}^{EC}$ / (m/s) | 24.15 | 24.83 | 23.25 | 23.40 | 17.03 | 11.07 |
| $v_{x,t_c}^{AC}$ / (m/s) | 18.66 | 19.17 | 20.76 | 20.91 | 21.49 | 21.52 |

where $t_c$ is the time of lane-change decision making, $\Delta s_{t_c}$ is the gap between EC and AC at the time $t_c$. $v_{x,t_c}^{EC}$ and $v_{x,t_c}^{AC}$ are the velocities of EC and AC at the time $t_c$. NE denotes Nash Equilibrium and SE denotes Stackelberg Equilibrium.

TABLE V
COST VALUES ON DRIVING PERFORMANCES OF THE EC IN SCENARIO A.

| Cost Values ($\times 10^5$) | Aggressive | | Normal | | Conservative | |
|---|---|---|---|---|---|---|
| | NE | SE | NE | SE | NE | SE |
| Driving Safety | 776 | 428 | 153 | 119 | 63 | 50 |
| Ride Comfort | 365 | 231 | 183 | 147 | 105 | 104 |
| Travel Efficiency | 114 | 79 | 275 | 215 | 394 | 329 |

where the cost values are the root mean square (RMS) values.

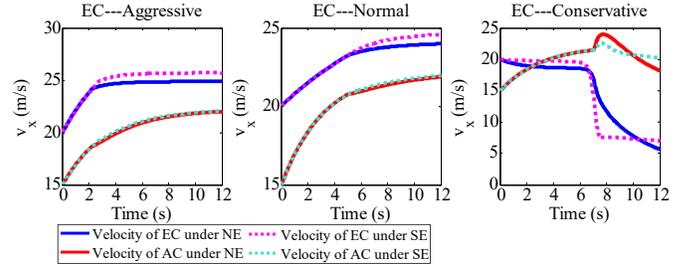

Fig. 8. Testing results of velocities in Scenario A.

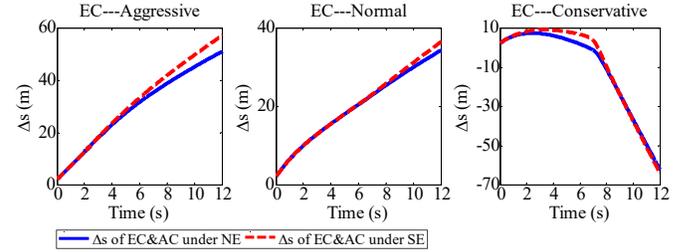

Fig. 9. Testing results of the gaps in Scenario A.

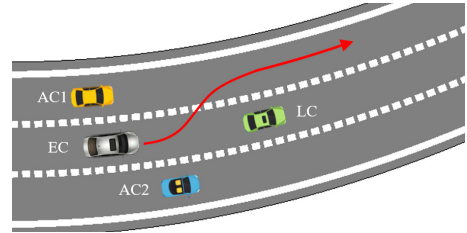

Fig. 10. Testing Scenario B.

### B. Scenario B

Scenario B is an overtaking maneuver on a curved highway with three lanes. As Fig. 10 shows, both LC and EC move on the middle lane. Since LC moves with a slow velocity, EC has to a make decision, slowing down to maintain a safe distance and following LC on the current lane or changing lanes to overtake. If EC chooses to change lanes, which side should be selected, the left side or the right side? AC1 and AC2 are moving on the left and right lanes. The behaviors of both AC1 and AC2 must be taken into consideration in the decision-making process of EC. At the initial moment, the longitudinal velocities of LC, EC, AC1 and AC2 are set as 15 m/s, 20 m/s, 15 m/s and 13 m/s. The initial position coordinates of LC, EC, AC1 and AC2 are (62, 0.95), (12, 0.04), (10, 4.03), and (15, -3.94). Figs. 11-13 show the testing results under different driving styles and different strategies.

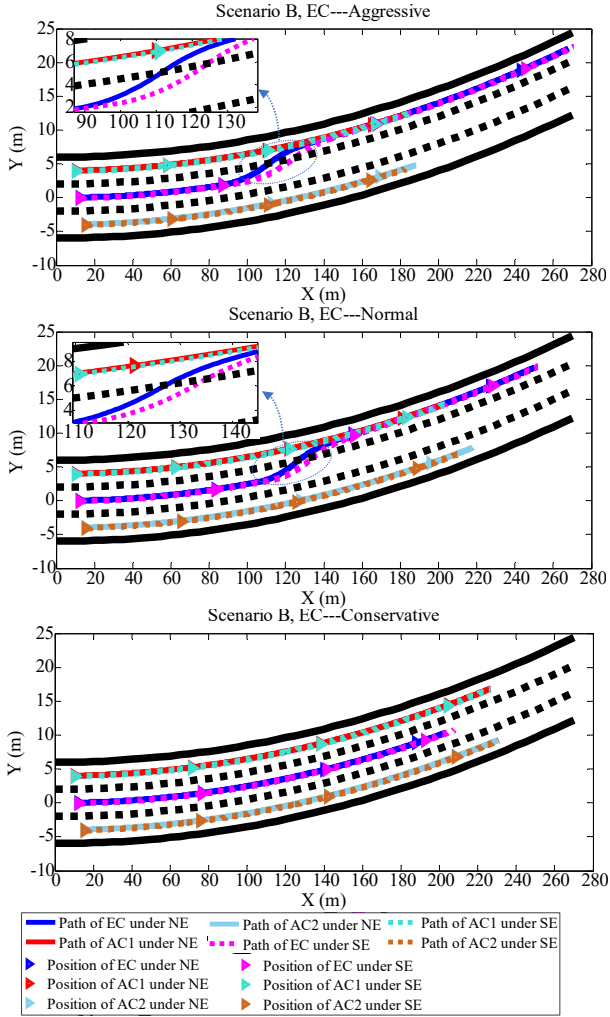

Fig. 11. Decision-making results in Scenario B.

It can be seen from Fig. 11 that the aggressive and normal modes make a left lane-change decision due to the smaller cost generated for the left lane. However, the conservative mode makes a lane keeping decision. The explanation for the above different decision-making results is that, the cost value of a lane-change for EC with a conservative style is larger than that of lane-keeping at this moment. In other words, under the conservative style, the EC cares more about driving safety rather than travel efficiency. The detailed decision-making results are listed in Table VI. It is found that, in order to improve the travel efficiency, the aggressive mode has a shorter decision-making time and a larger velocity for lane-change than the normal mode. Moreover, as reflected by the Table VII, the aggressive mode has the smallest cost value of travel efficiency, while the conservative mode has the smallest cost values of driving safety and ride comfort, which further validate the difference of preferred features in different driving styles. The similar conclusions also can be drawn from the results of Figs. 12 and 13. If EC's driving style is set as aggressive, to ensure travel efficiency, EC would speed up with a large acceleration and spend the shortest time finishing the overtaking process. If EC's driving style is normal, it would take more time to accelerate to complete the overtaking. However, if EC's driving style is set as conservative, it prefers to decelerate to keep a larger safe distance. Similar to Scenario A, it can be seen from Table VII that the cost values of the three driving performances in SE are smaller than those in NE, which indicates that SE provides better driving safety, ride comfort and travel efficiency for EC. In the normal driving style, the cost values of the driving safety, ride comfort and travel efficiency in SE are reduced by 31%, 34% and 27%, respectively.

TABLE VI
TESTING RESULTS OF DECISION MAKING IN SCENARIO B

| Testing Parameters | Aggressive | | Normal | | Conservative | |
|---|---|---|---|---|---|---|
| | NE | SE | NE | SE | NE | SE |
| $t_c$ / s | 2.98 | 3.44 | 3.90 | 4.20 | - | - |
| $\Delta s_{t_c}^1$ / m | 28.66 | 33.32 | 25.53 | 26.20 | - | - |
| $\Delta s_{t_c}^2$ / m | 26.22 | 31.21 | 22.55 | 24.06 | - | - |
| $v_{x,t_c}^{EC}$ / (m/s) | 27.03 | 27.22 | 23.43 | 23.44 | - | - |
| $v_{x,t_c}^{AC1}$ / (m/s) | 16.96 | 17.22 | 20.11 | 20.31 | - | - |
| $v_{x,t_c}^{AC2}$ / (m/s) | 16.23 | 16.52 | 19.82 | 20.04 | - | - |

where $t_c$ is the time of lane-change decision making, $\Delta s_{t_c}^i$ ($i=1,2$) denotes the gap between EC and ACi ($i=1,2$) at the time $t_c$. $v_{x,t_c}^{EC}$ and $v_{x,t_c}^{ACi}$ ($i=1,2$) denote the velocities of EC and ACi ($i=1,2$) at the time $t_c$.

TABLE VII
COST VALUES ON DRIVING PERFORMANCES OF EC IN SCENARIO B.

| Cost Values (× 10⁴) | Aggressive | | Normal | | Conservative | |
|---|---|---|---|---|---|---|
| | NE | SE | NE | SE | NE | SE |
| Driving Safety | 289 | 192 | 152 | 105 | 36 | 31 |
| Ride Comfort | 280 | 269 | 191 | 142 | 61 | 53 |
| Travel Efficiency | 119 | 107 | 279 | 205 | 371 | 344 |

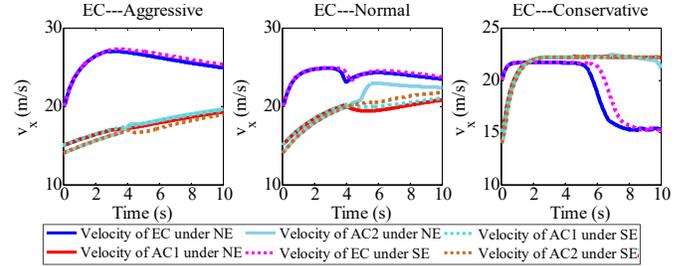

Fig. 12. Testing results of velocities in Scenario B.

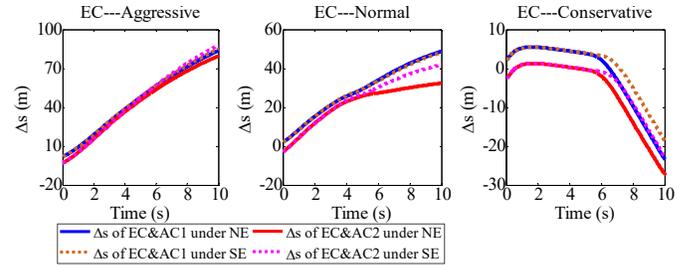

Fig. 13. Testing results of the gaps in Scenario B.

### C. Discussion of the Testing Results

Based on the above testing results under different scenarios, we can see that different driving styles would lead to different decision-making results for realizing the human-like concept of AV. The aggressive mode cares more about travel efficiency, and the conservative mode gives the priority to driving safety and ride comfort, while the normal mode tend to make a trade-off among travel efficiency, ride comfort and driving safety. Additionally, the two noncooperative game theoretic approaches, which are used to simulate the interaction and decision-making behaviors of human drivers, could help AVs make reasonable and proper decisions. Compared with NE, the SE shows better performance in terms of the driving safety, ride comfort and travel efficiency.

In addition, comparing to machine learning algorithms, although the proposed method shows a great ability to address decision-making problem of AVs, it still has limitations to handle various complex scenarios. When the traffic scenario and conditions vary, the models required for decision making would be far different. Thus, a unified, high-efficiency, and generic modeling framework is worthwhile exploring to solve this issue in the future. Additionally, with the increase of the system complexity, a larger amount of computation resources and a more powerful execution capability for decision-making are required. Therefore, further improvement of the computation efficiency of the algorithm is another direction of the future work.

## VII. CONCLUSIONS AND FUTURE WORKS

This paper presents a human-like decision making framework for AVs. Different driving styles associated with driving safety, ride comfort and travel efficiency are defined for AVs. Based on the decision-making model considering the different driving styles and social interaction characteristics, Nash equilibrium and Stackelberg game theory are utilized to address the noncooperative decision making problem. In addition, the potential field method and MPC are used to plan the collision-avoidance path and provide predicted motion states for the human-like decision making module. Finally, two testing scenarios are established to evaluate the performance of the decision-making framework. The testing results indicates that the developed framework is able to make reasonable decisions under different driving styles, which is human-like and can provide personalized choices for different passengers.

Our future work will focus on the algorithm improvement, in terms of its computation efficiency and the adaptivity to various complex traffic scenarios. Besides, the hardware-in-the-loop experiments will be conducted to validate the performance of the proposed algorithms in real time.

REFERENCES

[1] Y. Huang, H. Wang, A. Khajepour, H. Ding, K. Yuan, and Y. Qin, "A novel local motion planning framework for autonomous vehicles based on resistance network and model predictive control," *IEEE Trans. Veh. Technol.*, vol. 69, no.1, pp. 55–66, Jan. 2020.
[2] F. Fabiani, and S. Grammatico, "Multi-vehicle automated driving as a generalized mixed-integer potential game," *IEEE Trans. Intell. Transp. Syst.*, vol. 21, no. 3, pp. 1064–1073, Mar. 2020.
[3] R. Schubert, "Evaluating the utility of driving: Toward automated decision making under uncertainty," *IEEE Trans. Intell. Transp. Syst.*, vol. 13, no. 1, pp. 354–364, Mar. 2012.
[4] S. Noh, "Decision-making framework for autonomous driving at road intersections: Safeguarding against collision, overly conservative behavior, and violation vehicles," *IEEE Trans Ind Electron*, vol. 66, no. 4, pp. 3275–3286, Apr. 2018.
[5] C Yu, X Wang, X Xu, M. Zhang, H. Ge, J. Ren, L. Sun, B. Chen, and G. Tan, "Distributed multiagent coordinated learning for autonomous driving in highways based on dynamic coordination graphs," *IEEE Trans. Intell. Transp. Syst.*, vol. 21, no. 2, pp. 735–748, Feb. 2020.
[6] H. Bai, S. Cai, N. Ye, D. Hsu, and W. S. Lee, "Intention-aware online POMDP planning for autonomous driving in a crowd," in *Proc IEEE Int Conf Rob Autom.*, Seattle, WA, United States, May. 2015, pp. 454–460.
[7] T. Rosenstatter, and C. Englund, "Modeling the level of trust in a cooperative automated vehicle control system," *IEEE Trans. Intell. Transp. Syst.*, vol. 19, no. 4, pp. 1237–1247, Apr. 2018.
[8] A. Furda and L. Vlacic, "Enabling safe autonomous driving in real-world city traffic using multiple criteria decision making," *IEEE Intell. Transp. Syst. Mag.*, vol. 3, no. 1, pp. 4–17, Apr. 2011.
[9] J. Chen, P. Zhao, H. Liang, and T. Mei, "A multiple attribute-based decision making model for autonomous vehicle in urban environment," in *Proc IEEE Intell Veh Symp*, Dearborn, MI, United States, June. 2014, pp. 480–485.
[10] X. Wu, X. Xu, X. Li, K. Li, and B. Jiang, "A kernel-based extreme learning modeling method for speed decision making of autonomous land vehicles," in *2017 6th Data Driven Control and Learning Systems (DDCLS)*, Chongqing, China, May. 2017, pp. 769–775.
[11] Y. Liu, X. Wang, L. Li, S. Cheng, and Z. Chen, "A novel lane change decision-making model of autonomous vehicle based on support vector machine," *IEEE Access*, vol. 7, pp. 26543–26550, 2019.
[12] L. Li, K. Ota, and M. Dong, "Humanlike driving: Empirical decision-making system for autonomous vehicles," *IEEE Trans. Veh. Technol.*, vol. 67, no. 8, pp. 6814–6823, Aug. 2018.
[13] C. You, J. Lu, D. Filev, and P. Tsiotras, "Highway traffic modeling and decision making for autonomous vehicle using reinforcement learning," in *Proc. IEEE Intell. Veh. Symp.*, Changshu, China, June. 2018, pp. 1227–1232.
[14] X. Xu, L. Zuo, X. Li, L. Qian, J. Ren, and Z. Sun, "A reinforcement learning approach to autonomous decision making of intelligent vehicles on highways," *IEEE Trans. Syst. Man Cybern. Syst.*, to be published, Doi: 10.1109/TSMC.2018.2870983
[15] D. C. K. Ngai and N. H. C. Yung, "A multiple-goal reinforcement learning method for complex vehicle overtaking maneuvers," *IEEE Trans. Intell. Transp. Syst.*, vol. 12, no. 2, pp. 509–522, Jun. 2011.
[16] J. Wei and J. M. Dolan, "A robust autonomous freeway driving algorithm," in *Proc. IEEE Intell. Veh. Symp.*, Xi'an, China, June. 2009, pp. 1015–1020.
[17] J. Pérez, V. Milanés, E. Onieva, J. Godoy, and J. Alonso, "Longitudinal fuzzy control for autonomous overtaking," in Proc. IEEE Int. Conf. Mechatronics, Istanbul, Turkey, Apr. 2011, pp. 188–193.
[18] A. Li, H. Jiang, J. Zhou, and X. Zhou, "Learning human-like trajectory planning on urban two-lane curved roads from experienced drivers," *IEEE Access*, Vol. 7, pp. 65828–65838, May. 2019.
[19] M. Lindorfer, C. F. Mecklenbraeuker, and G. Ostermayer, "Modeling the imperfect driver: Incorporating human factors in a microscopic traffic model," *IEEE Trans. Intell. Transp. Syst.*, vol. 19, no. 9, pp. 2856–2870, Sept. 2018.
[20] S. Moon, and K. Yi, "Human driving data-based design of a vehicle adaptive cruise control algorithm," *Vehicle System Dynamics*, vol. 46, no. 8, pp. 661–690, Jun. 2008.
[21] A. Li, H. Jiang, Z. Li, J. Zhou, and X. Zhou, "Human-like trajectory planning on curved road: learning from human drivers," *IEEE Trans. Intell. Transp. Syst.*, DOI: 10.1109/TITS.2019.2926647, July. 2019.
[22] X. Gu, Y. Han, and J. Yu, "A Novel Lane-Changing Decision Model for Autonomous Vehicles Based on Deep Autoencoder Network and XGBoost," *IEEE Access*, vol. 8, pp. 9846–9863, Jan. 2020.
[23] C. W. Chang, C. Lv, H. Wang, D. Cao, E. Velenis, and F. Y. Wang, "Multi-point turn decision making framework for human-like automated driving," in *2017 IEEE 20th International Conference on Intelligent Transportation Systems (ITSC)*, DOI: 10.1109/ITSC.2017.8317831, Oct. 2017.
[24] P. De Beaucorps, T. Streubel, A. Verroust-Blondet, F. Nashashibi, B Bradai, and P. Resende. "Decision-making for automated vehicles at intersections adapting human-like behavior," in *2017 IEEE Intelligent Vehicles Symposium (IV)*, pp. 212–217, Jun. 2017.
[25] C. Guo, K. Kidono, R. Terashima, and Y. Kojima, "Toward human-like behavior generation in urban environment based on Markov decision process with hybrid potential maps," in *2018 IEEE Intelligent Vehicles Symposium (IV)*, pp. 2209–2215, Jun. 2018.
[26] H. Yu, H. E. Tseng, and R. Langari, "A human-like game theory-based controller for automatic lane changing," *Transp. Res. Part C Emerg. Technol.*, vol. 88, pp. 140–158, Mar. 2018.
[27] M. Wang, S. P. Hoogendoorn, W. Daamen, B. van Arem, and R. Happee, "Game theoretic approach for predictive lane-changing and car-

following control," *Transp. Res. Part C Emerg. Technol.*, vol. 58, pp. 73-92, Sept. 2015.
[28] Y. Xing, C. Lv, D. Cao. Personalized vehicle trajectory prediction based on joint time-series modeling for connected vehicles. IEEE Transactions on Vehicular Technology, 69(2), pp.1341-1352, 2019.
[29] J. Wang, G. Zhang, R. Wang, S. C. Schnelle, and J. Wang, "A gain-scheduling driver assistance trajectory-following algorithm considering different driver steering characteristics," *IEEE Trans. Intell. Transp. Syst.*, vol. 18, no. 5, pp. 1097–1108, May. 2017.
[30] C. Lv, X. Hu, A. Sangiovanni-Vincentelli, Y. Li, C. M. Martinez, and D. Cao, "Driving-style-based codesign optimization of an automated electric vehicle: a cyber-physical system approach," *IEEE Trans. Ind. Electron.*, vol. 66, no. 4, pp. 2965–2975, Apr. 2018.
[31] C. M. Martinez, M. Heucke, F. Y. Wang, B. Gao, and D. Cao, "Driving style recognition for intelligent vehicle control and advanced driver assistance: A survey," *IEEE Trans. Intell. Transp. Syst.*, vol. 19, no. 3, pp. 666–676, Aug. 2017.
[32] Y. Xing, C. Lv, D. Cao, and C. Lu, "Energy oriented driving behavior analysis and personalized prediction of vehicle states with joint time series modeling," *Applied Energy*, vol. 261, pp. 114471, Mar. 2020.
[33] J. Wang, M. Dai, G. Yin, N. Chen, "Output-feedback robust control for vehicle path tracking considering different human drivers' characteristics," *Mechatronics*, vol. 50, pp. 402–412, Apr. 2018.
[34] K. Zhang, J. Wang, N. Chen, and G. Yin, "A non-cooperative vehicle-to-vehicle trajectory-planning algorithm with consideration of driver's characteristics," *Proc. Inst. Mech. Eng. Part D J. Automob. Eng.*, vol. 233, no. 10, pp. 2405–2420, July. 2018.
[35] Y. W. Chai, Y. Abe, Y. Kano, and M. Aeb, "A study on adaptation of SBW parameters to individual driver's steer characteristics for improved driver-vehicle system performance," *Veh Syst Dyn*, vol. 44, no. sup1, pp. 874–882, 2006.
[36] J. Wang, J. Wang, R. Wang, and C. Hu, "A framework of vehicle trajectory replanning in lane exchanging with considerations of driver characteristics," *IEEE Trans. Veh. Technol.*, vol. 66, no. 5, pp. 3583–3596, May. 2017.
[37] B. Zhou, Y. Wang, G. Yu, and X. Wu, "A lane-change trajectory model from drivers' vision view," *Transp. Res. Part C Emerg. Technol.*, vol. 85, pp. 609–627, Dec. 2017.
[38] B. Shi, L. Xu, J. Hu, Y. Tang, H. Jiang, W. Meng, and H. Liu, "Evaluating driving styles by normalizing driving behavior based on personalized driver modeling," *IEEE Trans. Syst. Man Cybern. Syst.*, vol. 45, no. 12, pp. 1502–1508, Apr. 2015.
[39] P. Hang, and X. Chen, "Integrated chassis control algorithm design for path tracking based on four-wheel steering and direct yaw-moment control," *Proc Inst Mech Eng Part I J Syst Control Eng*, vol. 233, no. 6, pp. 625–641, Jul. 2019.
[40] P. Hang, X. Chen, and F. Luo, "LPV/H∞ controller design for path tracking of autonomous ground vehicles through four-wheel steering and direct yaw-moment control," *Int. J. Automot. Technol.*, vol.20, no. 4, pp. 679–691, Aug. 2019.
[41] X. Na, and D. J. Cole, "Game-theoretic modeling of the steering interaction between a human driver and a vehicle collision avoidance controller," *IEEE Trans. Hum.-Mach. Syst.*, vol. 45, no. 1, pp. 25–38, Feb. 2015.
[42] X. Na, and D. J. Cole, "Application of open-loop stackelberg equilibrium to modeling a driver's interaction with vehicle active steering control in obstacle avoidance," *IEEE Trans. Hum.-Mach. Syst.*, vol. 47, no. 5, pp. 673–685, May. 2017.
[43] Q. Tu, H. Chen, and J. Li, "A potential field based lateral planning method for autonomous vehicles," *SAE Int. J. Passeng. Cars - Electron. Electr. Syst.*, vol. 10, no. 1, pp. 24–34. Sept. 2016.